# Measuring Corporate Human Capital Disclosures: Lexicon, Data, Code, and Research Opportunities[*]

Elizabeth Demers[$]
Victor Xiaoqi Wang[€]
Kean Wu[£]

February 2024

## Abstract

Human capital (HC) is increasingly important to corporate value creation. Unlike other assets, however, HC is not currently subject to well-defined measurement or disclosure rules. We use a machine learning algorithm (word2vec) trained on a confirmed set of HC disclosures to develop a comprehensive list of HC-related keywords classified into five subcategories (DEI; health and safety; labor relations and culture; compensation and benefits; and demographics and other) that capture the multidimensional nature of HC management. We share our lexicon, corporate HC disclosures, and the Python code used to develop the lexicon, and we provide detailed examples of using our data and code, including for fine-tuning a BERT model. Researchers can use our HC lexicon (or modify the code to capture another construct of interest) with their samples of corporate communications to address pertinent HC questions. We close with a discussion of future research opportunities related to HC management and disclosure.

Data Availability: Data are available from the public sources cited in the text.

Supplemental material can be accessed by clicking the link in the Appendix.

---

[*] We thank Bruce Dehning (the Editor) and two anonymous reviewers for their helpful comments and suggestions.
[$] School of Accounting and Finance, University of Waterloo, Ontario, Canada. elizabeth.demers@uwaterloo.ca.
[€] College of Business, California State University, Long Beach, CA, USA. victor.wang@csulb.edu (corresponding author).
[£] Saunders College of Business, Rochester Institute of Technology, Rochester, NY, USA. kwu@saunders.rit.edu.

# Measuring Corporate Human Capital Disclosures: Lexicon, Data, Code, and Research Opportunities

## I. INTRODUCTION

In the modern, knowledge-based economy, human capital is often considered to be a company's most important asset (Batish et al. 2021) and a key driver of firm value (e.g., Edmans 2011), even while sustainable investing and investor awareness of social justice issues are also on the rise. Investor demand for information about how firms manage their human capital (HC) has therefore become correspondingly important.[1] However, measuring textual HC disclosures is not an easy task. Human capital management (HCM) is multi-faceted, including aspects such as recruitment, talent development, health and safety, compensation management, and fostering an inclusive environment. Prior studies related to HC disclosures rely on short lists of keywords or manual coding schemes that are often not adequate for capturing the multi-dimensionality of HCM practices, nor are they suitable for large-sample analyses. In this paper, we develop a comprehensive list of HC-related keywords by using a semi-supervised machine learning algorithm, the *word2vec* model (Mikolov, Chen, Corrado, and Dean 2013). Researchers can use our lexicon to identify and measure HC disclosures with a view to examining many timely research questions related to corporate HC management.

The first and most important step involved in quantifying textual HC disclosures is to reliably identify HC-related contents. Before the SEC mandated the human capital (HC) disclosures that are now provided in a separate section within Item 1 of Form 10-Ks, firms voluntarily provided HC-related disclosures in various documents or venues such as regulatory filings, sustainability reports, earnings conference calls, and company websites. Moreover, such disclosures were often dispersed and embedded in other non-HC-specific content. We capitalize on the SEC's new disclosure requirements to train our model on these now reliably identifiable HC-specific disclosures. To begin, we hand collect HC disclosures from 10-Ks filed by approximately 4,000 firms during the first year of the new regulation, from November 2020 to November 2021. Training our model on this corpus of unambiguously HC-related documents consisting of almost two million words ensures that our lexicon has high construct validity. Our lexicon covers five broad topics in HCM, including "Diversity, Equity, and Inclusion (DEI)" (253 terms), "Health and Safety" (227 terms), "Labor Relations and Culture" (362 terms), "Compensation and Benefits (283 terms)," and "Demographics and Other" (160 terms). For "Health and Safety," we further classify the keywords into those that are specific to COVID-19 (70 terms) and those that are more generic (157 terms).

---

[1] For example, in their survey of institutional investors, Morrow Sodali (2019) found that 83 percent of respondents indicated that the ESG topic most in need of improved disclosures was human capital. See also discussions about the petition from the Working Group on Human Capital Accounting Disclosure to the SEC (Posner 2022).

With a view to contributing to recent initiatives that promote the use of novel datasets to answer interesting research questions (e.g., Hoitash and Hoitash 2022; Wang and Wang 2022; Abu-Khadra and Olsen 2023), we share our lexicon, together with the approximately 4,000 machine-readable corporate textual disclosures and Python code used to develop the keyword list. We also provide sample code that can be used together with the word list to extract HC measures from any other user-provided collection of textual documents. Researchers can use our lexicon and/or adapt our codes to create alternative HC measures that suit their particular research setting. Using our code, researchers can also apply the *word2vec* algorithm to develop word lists for other (i.e., non-HC) constructs of interest by supplying their own list of initial seed words to the algorithm. The *word2vec* algorithm is a promising new technique in textual analysis as it overcomes a major weakness of the traditional dictionary-based approach, which often relies on a non-comprehensive list of keywords identified by the researcher. The latter may be based upon the researcher's own limited perceptions, training, and experiences, or the selection of source documents (e.g., textbooks), and therefore may not capture the range of terminology in current use by a varied set of corporate reporters. Our approach overcomes these limitations.

Recent developments in machine learning provide alternative ways to identify and classify text, such as using transformer-based Large Language Models (LLMs) like BERT (Devlin, Chang, Lee, and Toutanova 2019). However, the application of such models to a specific domain often necessitates pre-training a new model on a large corpus of domain-specific text and/or fine-tuning the pre-trained base model using labeled datasets. Two notable examples are FinBERT (Huang, Wang, and Yang 2023), which is pre-trained on finance text, and ClimateBERT (Webersinke, Kraus, Bingler, and Leippold 2022), which is pre-trained on climate disclosure text. The HC disclosures that we collect and share with this study can serve as valuable resources for pre-training a BERT model specific to HC disclosures or for fine-tuning the base BERT model for HC disclosure classification. In addition, our keyword list can also serve as a useful tool for facilitating pre-training or fine-tuning by identifying HC-specific documents or pinpointing human-labeled samples that may require additional scrutiny. To illustrate the potential use of our data for training or fine-tuning a domain-specific LLM, we fine-tune the base BERT model for identification of HC content using the confirmed HC disclosures that we collected as labeled datasets.

In the remainder of the paper, we begin by providing a review of the prior HC literature, with a special focus on the measurement of HCM practices and HC disclosures. This review reveals that the literature lacks a comprehensive measure for quantifying textual HC disclosures or otherwise gauging a firm's HCM practices for a large cross-section of firms. We then elaborate on how we develop our lexicon as a new alternative for measuring corporate HC disclosures and HCM practices, and demonstrate how to use the lexicon by constructing HC disclosure measures for a large collection of proxy statements containing more than 3.3 billion words. We further illustrate an alternative way of using our data to classify HC disclosures by fine-tuning the base BERT model, and discuss the advantages and disadvantages of each of keyword-based and machine-learning-based classifications. Finally, we propose a number of avenues for future HC research that could be pursued using our lexicon and our data.

## II. LITERATURE REVIEW

Human capital is a topic of interest to researchers in many disciplines, including management, strategy, and organizational behavior, amongst others. Within the realm of accounting and finance, two streams of literature are noteworthy: i) the implications of HCM practices for the value creation process; and ii) corporate disclosure decisions related to their HCM. These two streams of literature are interconnected since firm disclosures offer an important channel for outsiders to understand a firm's HCM practices and their implications for firm value.

In the context of value creation, prior studies find that various aspects of HCM practices contribute to firm value, including compensation (Rayton 2003), employee satisfaction (Edmans 2011; Green, Huang, Wen, and Zhou 2019), and training (Molina and Ortega 2003). Because corporate disclosures of HCM practices are scant and are often in a format that makes them hard to collect and process (in the rare instance where they were available prior to the new regulation), most previous studies rely on external sources of information to gauge a firm's HCM practices, including "best employer" ratings (e.g., Edmans (2011)) or crowd-sourced employee ratings from Glassdoor (e.g., Green et al. (2019)). Other researchers overcome the challenge imposed by lack of tabulated disclosures by using survey-based measures, although such studies are necessarily limited to the firms that respond to the survey, often resulting in response bias and small sample data analysis constraints.[2]

The second stream of HC literature examines voluntary corporate disclosure practices. These studies largely use disclosure indexes that are manually constructed by researchers who read the disclosure text and assign a value of *one* or *zero* to indicate the presence or absence of a particular disclosure item. This traditional method of manual content analysis suffers from several weaknesses, including that the disclosure items of which the indices are composed are subjectively chosen, the manual coding of these items is time-consuming, and as a result, the methodology is typically only useful for small sample studies and produces results that are often hard to replicate.

For example, Vithana, Soobaroyen, and Ntim (2021) create disclosure indexes for UK FTSE 100 firms according to a list of 22 disclosure items that include "Employee health and safety," "Employee diversity and equity," and "Employee involvement and engagement," among others. These authors find that HC disclosures are inadequate in both breadth and depth for the largest FTSE firms, but that firms embracing an employee relation ideology of unitarism tend to provide a higher level of disclosure. Another study by McCracken, McIvor, Treacy, and Wall (2018) uses a sample of FTSE 100 firms to measure the level of HC disclosure using a list of items covering four broad categories, including "Knowledge, Skills, and Abilities (KSA)," "Human Resource Development (HRD)," "Employee welfare," and "Organizational justice and equity." Each of these four categories consists of a slightly expanded list of terms. For example, HRD includes "apprenticeships," "career development," "graduates," "internships," "succession planning," "talent management and training," and other similar terms. They find that the top 100 UK companies significantly increased their level of HC disclosure after the UK Companies Act 2006 was amended in 2013 to require companies to report on their human capital.

---

[2] Bernstein and Beeferman (2015) provide a summary of this literature.

Using the Canadian setting, Cormier, Aerts, Ledoux, and Magnan (2009) examine HC disclosures on the corporate websites of 155 non-financial firms included in the Toronto Stock Exchange S&P/TSX Index. They find that voluntary HC disclosure reduces information asymmetry and that better corporate governance leads to a higher level of disclosure. They measure HC disclosures using a coding instrument that includes 33 items such as "hiring," "training," "performance assessment," and "employee satisfaction."

Some studies examine HC disclosures in the broader context of a firm's sustainability reporting. For example, Ehnert, Parsa, Roper, Wagner, and Muller-Camen (2016) use manual coding to examine the extent of sustainable Human Resource Management (HRM) reporting for Forbes Top 250 global companies by using the framework of the Global Reporting Initiative (GRI). This framework covers quantitative and qualitative factors such as employee demographics, employee turnover, labor relations, occupational health and safety, training and education, and diversity and equal opportunity. Somewhat surprisingly, they find that these large firms provide disclosures on HRM performance that are commensurate with those related to environmental performance, and that international differences in HRM disclosures are not as large as expected.

A few studies examine corporate disclosures related to intellectual capital or property, which includes human capital. For example, Bontis (2003) examines the intellectual capital disclosures of Canadian companies and finds that the propensity for disclosure is surprisingly low during the sample period; only 74 instances of disclosure were found among 10,000 annual reports. He uses a list of 39 terms, about one third of which relate to human capital (e.g., "human capital,"

"human assets," and "employee skill"). In a related very small sample study, Sujan and Abeysekera (2007) examine the intellectual property reporting practices of the 20 largest Australian firms by market capitalization. Using traditional manual content analysis with a coding scheme, they find that HC disclosures in annual reports were mostly qualitative, and that such disclosures lacked consistency across firms even though there was an observable trend towards more quantitative disclosures.

The SEC's 2020 mandate on HC disclosures has generated tremendous interest in this topic, as numerous recent studies examine mandatory or voluntary HC disclosures in the U.S. setting. For example, Batish et al. (2021) manually read the HC disclosures provided by the 100 largest firms under the new regulation and subjectively conclude that these early disclosures are too generic to be useful to investors. Zhang (2022) examines the determinants and consequences of voluntary HC disclosures by using machine learning to extract HC disclosures from annual reports in a period before the new SEC regulation took effect. Demers, Wang, and Wu (2024) provide a comprehensive analysis of all available 10-K HC disclosures from the first two years of the new regulation. They find that the attributes of these disclosures (e.g., length, specificity, and tone) vary significantly across firms, and that the disclosures overall do not appear to be sufficiently informative due to their lack of specificity and numerical intensity. Time trends suggest that there are learning effects (e.g., later filers appear to learn from earlier filers, and firms update their filings in the second year of the regulation), but also that there is reversion towards the mean as firms with weaker disclosures (e.g., shorter, less specific, less numerically intense, or less readable) improve their disclosures while previously better disclosers regress on these characteristics. Arif, Yoon, and Zhang (2022) find that the amount of HC disclosure in 10-K filings is negatively associated with contemporaneous employee turnover, suggesting that such disclosures provide new information about a company's human capital management practices. They also find that both equity markets and bond markets react to HC disclosures, although they do so differently, in ways that are probably explained by the respective investor groups' different pay-off functions. Using manual content analysis on a sample of S&P 500 firms, Michaelides and Vafeas (2023) find that Chief Human Resource Officers (CHROs) play a significant role in determining the quality of HC disclosures, especially when the CHROs are more powerful. However, CHROs have a smaller impact when they come from underrepresented groups, such as when they are women or ethnic minorities.

A few recent studies focus on particular aspects of HC disclosure such as gender diversity (Liang, Lourie, Nekrasov, and Yoo 2022), attraction and retention (Haslag, Sensoy, and White 2022), COVID-19 related issues (Mayew and Zhang 2022), and quantitative metrics (Bourveau, Chowdhury, Le, and Rouen 2022). Liang et al. (2022) obtain their measures of gender diversity from external data providers (Bloomberg and Revelio Labs). They find that companies with a higher percentage of female employees are more likely to voluntarily disclose their employees' gender diversity. Haslag et al. (2022) identify employee attraction and retention disclosures in 10-K filings by requiring a sentence to contain at least one word from each of two researcher-constructed word lists. They find that firms adjust their HC disclosures in response to changes in their workforce. Mayew and Zhang (2022) identify COVID-19 related HC disclosures by searching a few pandemic-related keywords in the firm's HC disclosures. They find that the amount of such disclosures is positively related to employee satisfaction with firms' response to the pandemic. Bourveau et al. (2022) manually collect quantitative metrics of HCM practices from annual reports in a sample period surrounding the SEC's 2020 regulation and find that there is a significant increase in quantitative disclosures in the post-regulation period.

In summary, the literature related to the role of HC in the value creation process largely uses HC measures obtained from external sources to gauge the impact of a firm's HCM practices on firm performance. The literature related to corporate HC disclosure decisions, in turn, tends to use relatively short and non-comprehensive researcher-defined lists of select items or terms. These lists vary from study to study and this approach is not well-suited to large-sample analyses. Zhang (2022) is an exception to this in that she uses a machine learning approach that has the potential to create a list of comprehensive keywords for capturing various dimensions of a firm's HCM practices. Her model is trained on a broad set of mostly non-HC corporate textual disclosures, however, which can significantly reduce the construct validity of the resulting lexicon. Our approach overcomes all these limitations by applying machine learning to a large sample of recent HC-specific corporate disclosures to develop an extensive HC-related dictionary of contemporary relevance with high construct validity.

In the next section, we discuss how we construct our list of HC-related keywords by applying a machine learning algorithm to nearly 4,000 corporate 10-K HC disclosures filed during the first year of the 2020 SEC regulation.

## III. METHODOLOGY AND LEXICON

**Word2Vec**

We use *word2vec*, a word embedding model proposed by Mikolov et al. (2013), to construct a comprehensive list of HC keywords. Based on the assumption that words occurring in a similar context (i.e., surrounding words) tend to have similar meanings (Harris 1954), this algorithm uses a neural network to learn word associations from a large collection of text. By representing each word as a numeric vector that captures the semantic qualities of that word, it allows the relationship between words to be determined using metrics from vector operations such as cosine similarity. This means that one can identify synonyms of a word by finding words whose vector representations of neighboring words are similar to those of the focal word. With a list of initial words and a collection of documents as inputs, the algorithm can expand the list to include words in the documents that are synonymous with the initial words. This capability renders *word2vec* an ideal tool for constructing dictionaries to capture topics that are multi-faceted and/or evolving, and thus for which it is too difficult or too costly for experts to manually categorize all related words and phrases. Multiple studies in accounting and finance have used this promising tool for constructing dictionaries to measure various constructs, including the sentiment of financial text (Du, Huang, Wermers, and Wu 2021), corporate culture (Li, Mai, Shen, and Yan 2021), and *voluntary* HC disclosures (Zhang 2022).

As with any machine learning task, the quality of inputs is crucial for the quality of outputs (Geiger et al. 2021). In the case of *word2vec*, the model should be trained on text that is known to discuss the topic for which a dictionary is being developed in order to guarantee the lexicon's construct validity. We therefore train the model on a large collection of documents unambiguously related to HC disclosures, which is to say the actual HC disclosures made under the new regulation that were hand-collected from 3,953 unique firms' 10-K filings.

Table 1 summarizes the sample selection process for these HC disclosures. We start with all the 10-K Forms filed with the SEC's online filing system (i.e., EDGAR) during the first year of the 2020 regulation (i.e., from November 9, 2020 to November 8, 2021). From these 7,185 filings, we exclude 3,219 forms from filers that are not included in the Compustat and CRSP databases. Most of these filers are private companies, trusts, closed-end funds, or partnerships that are not required to comply with the regulation, or that do not have a significant number of employees. After this exclusion, there remain 3,966 10-Ks from filers that are included in EDGAR, Compustat, and CRSP. From these we exclude five filings belonging to fiscal year 2019, three filings from firms that outsource all of their personnel (i.e., according to Compustat they do not have any employees), and two filings that do not contain HC disclosures. Finally, we exclude three duplicate filings, keeping only the first 10-K in the case of multiple filings by the same company. After this process, 3,953 unique corporate filings remain. We use the HC disclosures from these filings to train the *word2vec* model. Using one disclosure from each company ensures that each company is given equal weight.

Taken together, the HC disclosures of these 3,953 unique firms consist of approximately two million words. We share these disclosures and their unique company identifiers in a single text file named "JIS_Data_HC_Disclosures.txt." The identifiers include CIKs, company names, filing dates, and fiscal reporting periods. We provide a utility function in "JIS_HC1_util.py" for extracting the texts and associated company identifiers from the single text file and saving this data to individual text files or to a combined CSV file. All data and code are available for download from the link provided in the Appendix.

Another important input into *word2vec* modelling is the list of initial seed words. We next describe how we have chosen these words.

**Initial seed words**

To ensure that our initial seed words cover various important dimensions of HCM practices, we first identify five broad topics according to the recommendations of standard setters and survey evidence. In its preliminary framework for HC disclosures, SASB identified three key HCM issues that are especially relevant to the long-term sustainability of a company: "Employee health and safety," "Labor practices," and "Employee diversity, inclusion, and engagement" (SASB 2020). Our first three categories address these issues. We include "Compensation and Benefits" as an additional category because compensation and benefits are the second and third most important drivers of employee satisfaction according to a survey conducted by the Society for Human Resource Management (SHRM) (Miller 2016), with the top driver being "respectful treatment of employees at all levels," which is already included in DEI. Our last category captures employee demographics and other HCM issues.

To come up with the initial word lists for these five categories, we take a two-pronged approach. First, we examine the issues or metrics that firms should disclose as recommended by standard setters, regulators, and various interest groups such as institutional investors.[3] Second, we manually read a large sample of HC disclosures in 10-Ks to identify the common words and phrases that firms use to discuss their HCM practices. In Table 2, we present our seed words for each of the five categories. Because we train the *word2vec* model on actual HC disclosures, we can choose a large number of words and phrases as seed words, and even include certain words that may appear to be too general in the context of a full 10-K document, but that are domain-specific in the context of HC disclosures. This allows us to identify words and phrases that otherwise would have been missed had a smaller number of seed words been used. For example, the words "fair" and "fairly" can take multiple different meanings in a 10-K; however, in the context of HC disclosures, they most likely refer to whether employees receive fair treatment or are treated fairly. Including words such as these that are too general in the larger context of a 10-K, but that are specific enough in the narrow context of HC disclosures, allows us to uncover other

---

[3] Zhang (2022) provides a comprehensive summary of these recommendations in her Online Appendix 1.

HC-related words that may be missed when the learning algorithm is applied to entire 10-Ks as in Zhang (2022).

**Final Word lists**

We feed the seed words into *word2vec* to allow the model to discover similar words and phrases. Words and phrases are considered to be "similar" if they are used in the same or similar context as the seed words. For these similar words and phrases, we initially keep those with a cosine similarity score of at least 0.5, rather than keeping the top-300 most similar words for each of the seven categories as in Zhang (2022) or the top-500 words for each of the five categories as in K. Li et al. (2021).[4] We make this choice for several reasons. First, keeping the top "n" words for each category assumes that there is an equal number of similar words for each category. This is not a valid assumption. Among the various dimensions of HCM, some dimensions can legitimately have more keywords than others due to their greater complexity. Second, within each category, choosing the top "n" words may give some seed words an unduly high weight simply because they happen to have more similar words with high cosine similarities.[5] This can reduce the comprehensiveness of the final word list. By choosing a threshold based on cosine similarity, we do not force each category to have the same initial weight and we do not allow certain seed words to dominate the list. Third, prior studies show that using a cosine similarity threshold of 0.5 yields stronger results. For example, Kee (2019) proposes a new approach for identifying peer firms using *word2vec* and finds the new approach outperforms existing industry classification systems when the threshold for the cosine similarity is greater than 0.5. Erfani, Cui, and Cavanaugh (2021) use *word2vec* to identify similar risks across different projects based on risk descriptions and find that two risks are meaningfully related to each other when the cosine similarity is above 0.5.

---

[4] With word embeddings, the cosine similarity score is bounded by negative one and positive one. A score of positive one indicates that two words always occur in the same context. A negative score indicates that the two words rarely occur in the same context and they are possibly antonyms.

[5] If seed word "A" has a large number (say, more than 300) of similar words with high similarity scores (say, all above 0.80), whereas seed word "B" has many words with similarity scores that are well above the reasonable 0.5 threshold but less than 0.8 (but some of them are very close to 0.8), then choosing only the top "300" words would exclude all the similar words of seed word "B" from being considered. This would result in an overrepresentation of words that are similar to seed word "A" and an underrepresentation of words that are similar to seed word "B."

To ensure that our word list captures antonyms, we also include words in our initial word list if they have a highly negative cosine similarity (i.e., cosine ≤ -0.5), consistent with the notion that even antonyms can appear in a similar context.[6] For example, companies may use "bias" or "anti-bias", two semantically opposite words, when they discuss their DEI policies. To investigate the extent of such words, in Figure 1 we plot the distribution of cosine similarity scores of all words with the seed words. The positively skewed distribution indicates that more words have a positive cosine similarity score, while the left tail of the distribution indicates that there is only a small percentage of words that have a high negative cosine similarity score.

At the cutoff of an absolute value of 0.5, our initial word list consists of 7,018 keywords. This large list of initial similar words ensures the comprehensiveness of our final word list. We take three steps to screen and classify these words. As the first step, we manually screen each of these keywords to make sure that they relate to HCM. Next, we assign a term to the category of the related seed word. When a term is similar to seed words from multiple categories (e.g., "DEI" and "Health and Safety"), we assign the term to the category for which the term has the highest average similarity score with the category's seed words. As the last step, we further review the classification of all the terms and manually re-classify them as appropriate. These procedures are standard with a *word2vec* implementation, and serve to improve the quality of the word lists. Both

---

[6] We thank an anonymous reviewer for this insightful suggestion.

K. Li et al. (2021) and Zhang (2022) screen out some words that do not pass the face validity check. Some of the words that we drop are not related to HCM, while others are related to HCM but may be too general for documents covering many topics other than HCM (e.g., "fair"). We exclude such words so that our final word list will have broader applicability to texts beyond the confirmed HC-specific disclosures used to construct our lexicon.

Table 3 presents the final word lists for each of the five categories in separate panels A through E. The keywords consist of seed words and similar words identified by the algorithm, for a total of 1,285 terms. Each of our word lists captures an important aspect of a firm's HCM practice, and collectively the lists enable the construction of measures that capture the firm's overall HC disclosure. Some of the keywords are acronyms. We provide their full spellings in Table 4. For the "Health and Safety" category, we further classify keywords into those related to COVID-19 (70 terms) and those that are more generic in nature (157 terms). The keywords related to COVID-19 are useful for examining questions related to firms' HC-related responses to the pandemic, but are less likely to be of interest to future researchers working outside of that particular context.

Following the prior literature, we classify the keywords into five categories largely based upon the keywords' similarity with the initial seed words for each category (i.e., computationally determined, but subject to a manual review as described above). To assess the appropriateness of this classification, we provide a visualization of the clusters of keywords that derive from our classification, and we compare the results to those generated by an automated classification using the K-Means clustering algorithm.[7] To visualize our classification, we extract the vectors of the

---

[7] We thank the anonymous reviewer for this suggestion.

keywords from the *word2vec* model, which represents each word using a vector of 100 numbers. We then reduce the dimensionality of the data using Principal Component Analysis (PCA) to three dimensions by keeping the first three components. Because we further break down keywords related to health and safety into those specific to COVID-19 and those that are more generic in nature, our classification consists of six categories (or clusters). For the automatic clustering, we use the K-Means clustering algorithm to classify these keywords into six clusters for easier comparison.

Panels A and B of Figure 2 present a 3-D visualization of our classification and the K-Means clustering classification, respectively. Each dot represents a keyword and the color of the dot represents its category or cluster. As shown in Panel A, the clustering of the keywords indicates that our classification works well. For example, the keywords for "DEI" and those for "Health and Safety (COVID)" cluster together coherently in the graph. In Panel B, the K-Means clustering may appear at first blush to be doing a better job. However, the largest K-Means cluster consists of approximately 80 percent of all keywords, whereas some of the K-Means clusters consist of a much smaller number of words. Manual inspection of the larger clusters suggests that many of the keywords could have been further classified into finer clusters. On the other hand, manual review of one of the smaller clusters shows that it represents a small subset of the DEI keywords included in our word list.

The previous comparison uses six clusters for K-Means clustering to correspond with the six clusters identified from our manual categorization. To alternatively find the *optimal* number of clusters, we calculate the silhouette scores for different numbers of clusters, ranging from N=2 to N=25. The theoretical value of the silhouette score (or coefficient) ranges from -1 to +1, with higher values indicating more coherent clusters. Figure 3 plots the silhouette scores against the number of clusters. As shown, the score peaks at four clusters, and it has the second highest value at six clusters. Given that most of the prominent HC disclosure frameworks (e.g., GRI 2016; ISO 2018; BlackRock 2022) identify more than four dimensions to corporate HCM, we consider the second peak, at six clusters, to be more appropriate for automated clustering. Investigating higher numbers of clusters, we observe, for example, that at 10 clusters, the largest cluster still consists of more than 70 percent of all keywords even though the smallest cluster contains only 14 keywords. At 20 clusters, the largest cluster still consists of approximately 70 percent of all keywords, whereas some of the smallest clusters contain fewer than 10 keywords. These additional analyses suggest that automated clustering has its limitations as this process is not able to distinguish the multidimensionality of human capital management to a satisfactory level. Finally, a further important insight offered by these analyses is that the algorithm's classification of most keywords into a single cluster indicates that our keyword list captures well the single broad underlying construct of human capital management.

The less-than-satisfactory results from automatic clustering could also be due to the fact that word embeddings generated by *word2vec* are context-independent. This means that a word has the same vector representation regardless of the context in which it is used. Due to this inherent limitation, *word2vec* is not able to effectively capture polysemy (i.e., words or phrases that can have different meanings depending on the context). Nonetheless, this negative impact should be relatively small and does not undermine the validity of our methodology because we train the *word2vec* model on a corpus of confirmed HC disclosures. The meanings of words used by companies to describe their HCM practices tend to be less varied across HC contexts compared to their use in general language. An alternative approach for refining the classification of HC disclosures into different topics is to use machine learning. We discuss more about this approach in the next section.

Having identified a comprehensive list of keywords covering various aspects of HCM, we next conduct two exploratory analyses of the HC disclosures from the first year of the SEC's 2020 regulation. For the first analysis, we explore how the HC disclosure topics vary across industries. As shown in Table 5, this analysis reveals notable patterns. For example, the "Oil, Gas, and Coal Extraction and Products" industry reports the most about employee health and safety, while the "Telephone and Television Transmission" industry reports the least. The high-risk nature of operations in the oil and gas sector necessitates more substantial communication about health and safety matters to stakeholders, which is reflected in their HC disclosures.

For the second analysis, we examine the temporal trends of HC disclosures during the first year based on the percentage of keywords for various topics relative to the total word count of HC disclosures. We calculate the average percentage of keywords for each successive 10-day period and plot the results in Figure 4. We do not observe notable patterns in the time series, probably because this analysis covers only one year and there are many factors that may affect companies' HC disclosures. We refer readers to Demers, Wang, and Wu (2024) for a thorough examination of how HC disclosures have evolved over time.

## IV. APPLICATIONS INVOLVING OUR HC LEXICON AND DATA

**An Application of Our Lexicon**

To illustrate the application of our lexicon, we use it to examine the trend of HC disclosures in proxy statements. We construct a few measures using our lexicon for HC disclosures in proxy statements over the period from 1994 (the starting year of electronic filing) to 2022. These measures include a count of HC keywords and the percentage of HC keywords relative to the total number of words in the proxy statement.[8] We construct measures for each of the five categories, as well as an aggregate measure that is computed as the sum of the five categorical measures. Technical details and the corresponding Python code underlying the measures' construction are provided in the Appendix.

In Figure 5, we plot the volume of HC disclosures in proxy statements from the period of 1994 to 2022. The left vertical axis shows the absolute HC keyword count and the right axis tracks the HC keyword count as a percentage of the total number of words in the proxy statement. The percentage of HC keywords was relatively stable over the 1994 to 2006 period. However, there was a sharp increase in 2007 due to a new SEC regulation that required firms to provide expanded disclosures about executive and director compensation. Our lexicon successfully captures this structural change in the reporting environment. In 2010 and 2011, there was a large increase in the absolute HC keyword count relative to 2009, which was due to the SEC's Proxy Disclosure Enhancements (effective February 28, 2010) that amended Regulation S-K Items 401, 402 and 407. Our lexicon again exhibits construct validity as it captures this structural change in regulatory reporting requirements.

In Figure 6, we break down the percentage of keywords by category. The category with the largest number of disclosures is "Compensation & Benefits," followed by "Demographics and Other," and "Labor Relations & Culture." This ranking is consistent with the contents of proxy statements, which provide a large amount of information about compensation, benefits and demographics, especially for executives and directors. Turning to the trend, there was an obvious jump in 2007 for disclosures in "Compensation & Benefits" and "Demographics & Other," due to the SEC's regulation change in that year. Even though the number of disclosures in the "DEI" and "Health & Safety" categories was low over the sample period, there was an evident increase in 2020 and 2021, consistent with both the growing awareness of the importance of DEI in recent years, as well as the many employee health and safety concerns that arose during the COVID-19 pandemic.

---

[8] Since this is for illustrative purposes, we make only a minor adaptation to our lexicon that involves excluding the "health and safety" keywords related to COVID-19 as these are obviously not relevant to the longer time series of documents used in our illustration.

In summary, this simple example illustrating the application of our lexicon to corporate proxy statements (i.e., "out-of-sample" documents) provides substantial evidence of the construct validity of our HC dictionary as the documented trends map into well-known regulatory and societal changes over the period under study.

**An Application Using Our Textual Data**

The advancement of Large Language Models (LLMs) underscores the growing importance of textual data, and particularly labeled textual data, because labeled textual data can be used to improve model performance through pre-training or fine-tuning. For example, users can leverage our HC disclosure text to fine-tune the base BERT model, thereby enabling BERT to identify HC disclosures and, even further, to classify the disclosures into specific topics. To showcase this potential, we use the confirmed HC disclosures underlying our study as an annotated dataset, supplemented by non-HC disclosures in Item 1 of 10-Ks. We extract 83,961 sentences from the HC disclosures and generate a random sample of twice as many non-HC sentences. Following the standard practice in machine learning, we use 80 percent of these sentences as the training set and 20 percent as the test set. We use "bert-base-uncased" as the base model and fine-tune it for a binary classification task (i.e., HC vs non-HC).[9] We fine-tune the model on a single powerful NVIDIA A100 GPU on Google Colab Pro.[10] Evaluation on the test set indicates an accuracy score of 97.46 percent, precision of 96.05 percent, recall of 96.33 percent, and F1 score of 96.19 percent. These results suggest that the fine-tuned model performs very well in out-of-sample prediction.

We next apply the fine-tuned model to 10-Q disclosures to see whether it continues to perform well on disclosure text that is different from the training set (i.e., contents in 10-Qs tend to be different from those in the 10-Ks that are the source of the training data). For this exercise, we manually label 1,000 sentences, 437 of which are confirmed HC sentences. We use the fine-tuned model to predict whether a sentence is related to HC, at a probability threshold that maximizes the F1 score.[11] At the probability threshold of 86 percent, the corresponding precision, recall, and F1 score are 85.16 percent, 85.35 percent, and 85.26 percent, respectively, for the HC class. These metrics indicate that the fine-tuned model performs impressively well on out-of-sample classification.

The preceding illustration demonstrates the use of our data for fine-tuning a BERT model for binary classification of HC disclosures. Further classification of HC content into more granular topics can also be done using a BERT model or another LLM. One way to approach this task is to first generate text embeddings for each HC sentence, and then to group similar sentences together using an automatic clustering algorithm.

**Keyword- Versus Machine-learning-based Classification**

---

[9] For a clear and accessible demonstration on how to fine-tune a BERT model, see Nichite (2022).
[10] The hyperparameters are: epochs = 2, batch_size = 16, learning_rate = 2e-05, and warmup_steps = 10000.
[11] For a guide on how to implement this automatic selection process, see Blancas (2022).

In the preceding subsections, we have illustrated how to use our lexicon and data for identifying and classifying HC disclosures using two different approaches, keyword-based vs machine-learning-based. Each of these methods has its own advantages and limitations, which we briefly discuss in this section.

Keyword-based classification may require pre-processing for normalizing certain words such as lemmatization and phrasing.[12] These additional steps may introduce noise, bias, or errors in the form of incorrect term reduction, loss of important contextual information, or incorrect parsing of complex phrases. Moreover, the keyword-based approach provides a deterministic result based upon the presence or absence of certain keywords and may not capture the context and other nuances of the language. Nonetheless, the keyword-based approach is transparent and easy to use. Such transparency allows fellow researchers to easily understand the classification process and also allows them to better evaluate, replicate, and potentially build on the prior research. The inherent simplicity of this approach makes it an accessible tool for researchers across various levels of expertise.

On the other hand, an ML-based approach, especially if it leverages the state-of-the-art LLMs, may provide better performance due to the ability of these models to understand complex linguistic and semantic features. However, ML models may need to be pre-trained on domain-specific text or fine-tuned on annotated datasets for improved performance. This requires a high level of technical expertise. The utilization of ML-based approaches may also be constrained by the availability of high-quality, domain-specific text or by the tremendous effort involved in creating a labeled dataset. Ultimately, the decision regarding which method to use should be made based upon the nature of the research question, the characteristics of the text under analysis, and any other constraints that the researcher may face, including their own technical expertise.

---

[12] An alternative approach is to employ regular expressions to account for various forms of a word or phrase. However, this method can become cumbersome, particularly when dealing with a large number of keywords.

## V. FUTURE RESEARCH OPPORTUNITIES

As the world economy continues to be predominantly knowledge-based and the social pillar of ESG becomes increasingly important (Drei et al. 2019; CPA Canada), there is a growing interest on the part of many stakeholder groups related to corporate HCM practices. In this section, we suggest some ideas that researchers in accounting and finance may pursue using our lexicon and data. As discussed in the literature review, prior studies examining the relation between HCM practices and firm value tend to use measures from external sources. There is a great opportunity to construct measures from firm-provided disclosures to gauge HCM practices for a large cross-section of firms. Using these measures as a complement to, or substitute for, existing measures, researchers may conduct more powerful tests on the relation between HCM practices and firm performance. Moreover, our lexicon covers five major categories, allowing for the construction of more granular measures of HCM practices. With these finer measures, researchers can study which employee treatment practices contribute more to value creation or to other important performance outcomes such as innovation. Answers to such questions have significant practical implications.

Firms can provide HC disclosures through many channels, including their annual reports, quarterly reports, proxy statements, earnings conference calls, and sustainability reports. Beattie and Smith (2010) surveyed HR directors of UK listed companies and found that managers consider annual reports to be the most effective venue for disclosing a firm's HC policy and practices to external parties. Since that time, however, there has been a dramatic increase in corporate issuances of stand-alone sustainability reports, and most companies offer a significant amount of sustainability-related information on their websites. There is a notable dearth of large-sample empirical evidence related to the determinants and consequences of companies' choice of HC disclosure channel(s). There are rich research opportunities to examine what, where, and why firms provide particular HC disclosures, and what the implications of these choices are to the firm's various stakeholders.

Among alternative disclosure channels, earnings conference calls are noteworthy. As a special form of corporate voluntary disclosure, conference calls offer an opportunity for interaction between corporate executives and financial analysts. The Q&A portion of conference calls, which is more extemporaneous (Chen, Demers, and Lev 2018), has been shown to elicit informative disclosures (Matsumoto, Pronk, and Roelofsen 2011). The extent to which managers volunteer, or analysts probe for, details about the firm's HCM is likely to be indicative of the materiality of HC to the firm's performance and prospects. Multiple studies develop measures using voluntary disclosures from earnings conference call transcripts. For example, Hassan, Hollander, van Lent, and Tahoun (2019) develop a firm-level political risk measure, Sautner, Van Lent, Vilkov, and Zhang (2023) provide measures of firms' climate change exposures, and K. Li et al. (2021) create a dictionary for measuring corporate culture from the Q&A portion of earnings calls. Future research could similarly develop a measure of the firm's HC-centricity or sensitivity, and explore how these measures influence capital market activity or analyst forecast updates, particularly in response to potentially HC-sensitive events such as social movements or changes in HC-related legislation.

Our word lists can be used to extract HC disclosures from conference call transcripts or other forms of corporate communications, and the extracted text can then form the basis from which other disclosure attributes such as readability (Li 2008), tone (Huang, Teoh, and Zhang 2014), specificity (Hope, Hu, and Lu 2016), numerical intensity (Henry 2008), forward-looking information (Muslu, Radhakrishnan, Subramanyam, and Lim 2014), and similarity (Brown and Tucker 2011) could be measured. These measures, in turn, can be used to gain a more nuanced understanding of a firm's HC disclosure decisions. For example, an overly optimistic tone that is not justified by real performance may indicate that a firm is "social washing." To the best of our knowledge, there is little evidence related to the extent, determinants, and consequences of social washing in HC disclosures, even though greenwashing in sustainability reporting has become an issue of top priority for regulators around the world.[13]

An additional opportunity available to researchers interested in the technical aspects or methodology of textual analysis is to pre-train or fine-tune a BERT model or another LLM on an extensive corpus of HC disclosure text. In our illustration, we have demonstrated this potential by obtaining satisfactory results through a modest fine-tuning process on a relatively small dataset, coupled with limited hyperparameter tuning. Those who plan to exploit this opportunity can incorporate our data into their training dataset and leverage our lexicon for identifying additional documents that are rich in HC-related content.

## VI. CONCLUSION

Human capital is frequently a company's most valuable asset, yet unlike other important asset classes, HC is not subject to well-defined measurement or disclosure rules. The construct of corporate human capital is itself so nebulous that the SEC actually refused to define it in the new disclosure regulations (Bourveau et al. 2022). We use a semi-supervised machine learning algorithm (*word2vec*) that is trained on a confirmed set of recent corporate HCM disclosures to develop an HC-related lexicon consisting of 1,285 terms classified into five disclosure categories: diversity, equity, and inclusion; health and safety; labor relations and culture; compensation and benefits; and demographics and other. We share our dictionary in a machine-readable format, together with the textual data and Python code used to construct it. We also provide an example of its application to corporate proxy statements. Researchers can modify our code to construct their own lexicon related to another topic of interest, or use our HC-related lexicon to pursue the many research opportunities that are available in this burgeoning area of study. In addition, we demonstrate the potential of using our textual data to fine-tune a LLM as an alternative approach for identifying and classifying HC disclosures.

---

[13] For example, the SEC launched the Climate and ESG Task Force within the Division of Enforcement in March 2021, with the mandate to identify and investigate potential corporate misconduct related to climate and ESG issues (SEC 2021).

**FIGURE 1 Distribution of Cosine Similarity Scores**

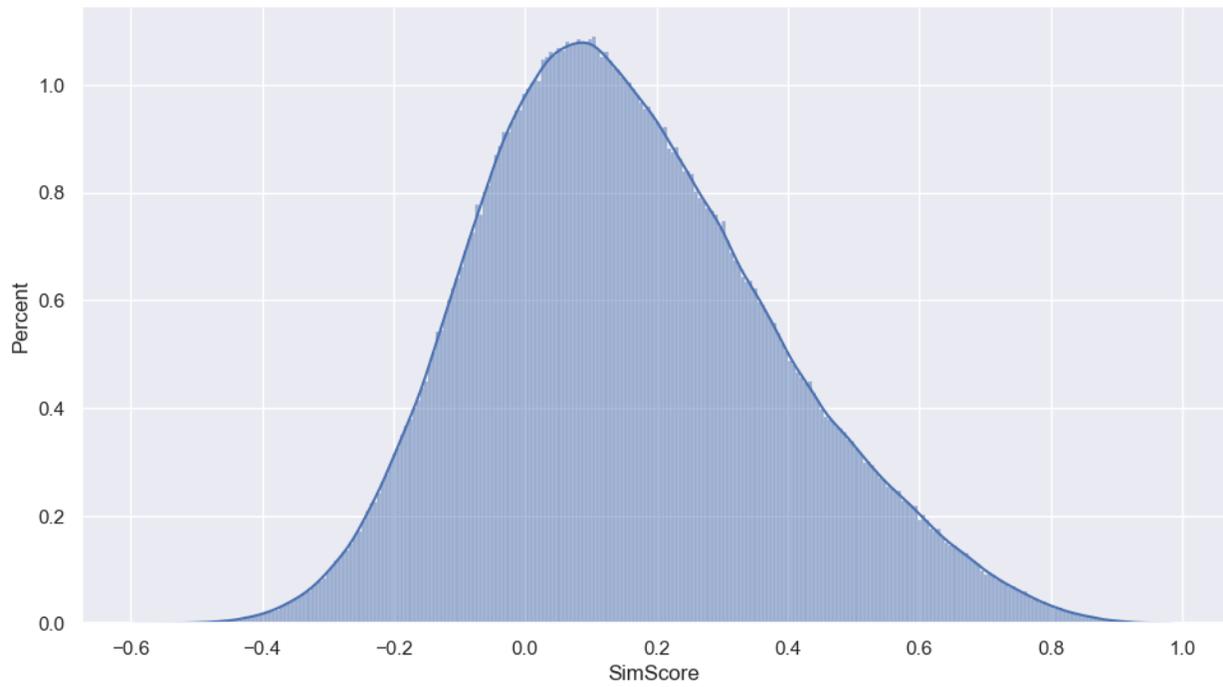

This graph plots the distribution of the cosine similarity scores between the seed words and other words in the corpus of HC disclosures used for training the *word2vec* model.

**FIGURE 2 Three-Dimensional Visualization of Keyword Classification**

**Panel A: Our classification**

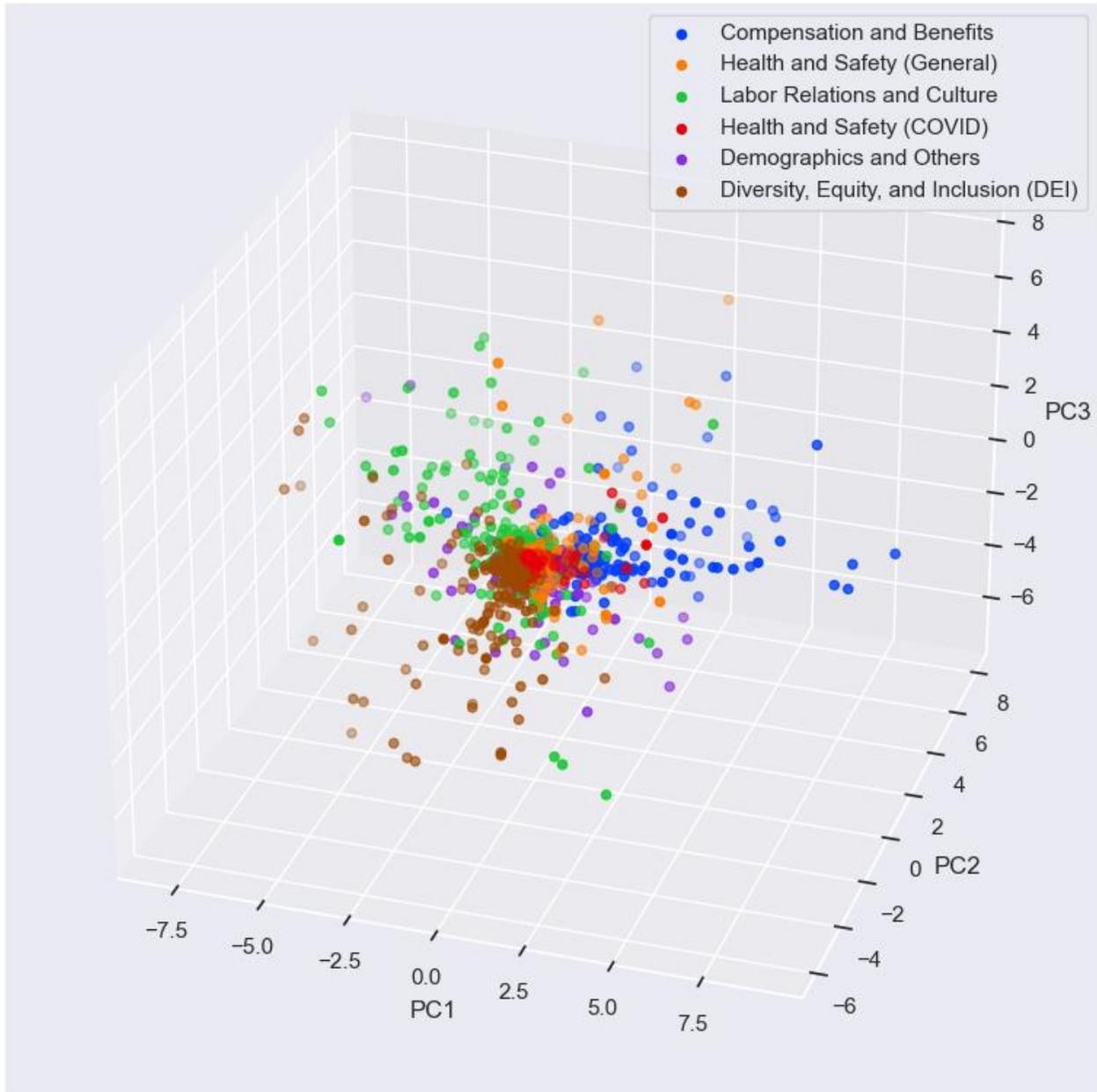

This graph visualizes our classification of keywords in a three-dimensional space. Each dot represents a keyword, with its color indicating the respective category of the keyword. The three axes capture the first three components from PCA analysis.

**FIGURE 2 Three-Dimensional Visualization of Keyword Classification**

**Panel B: Automated K-Means Clustering**

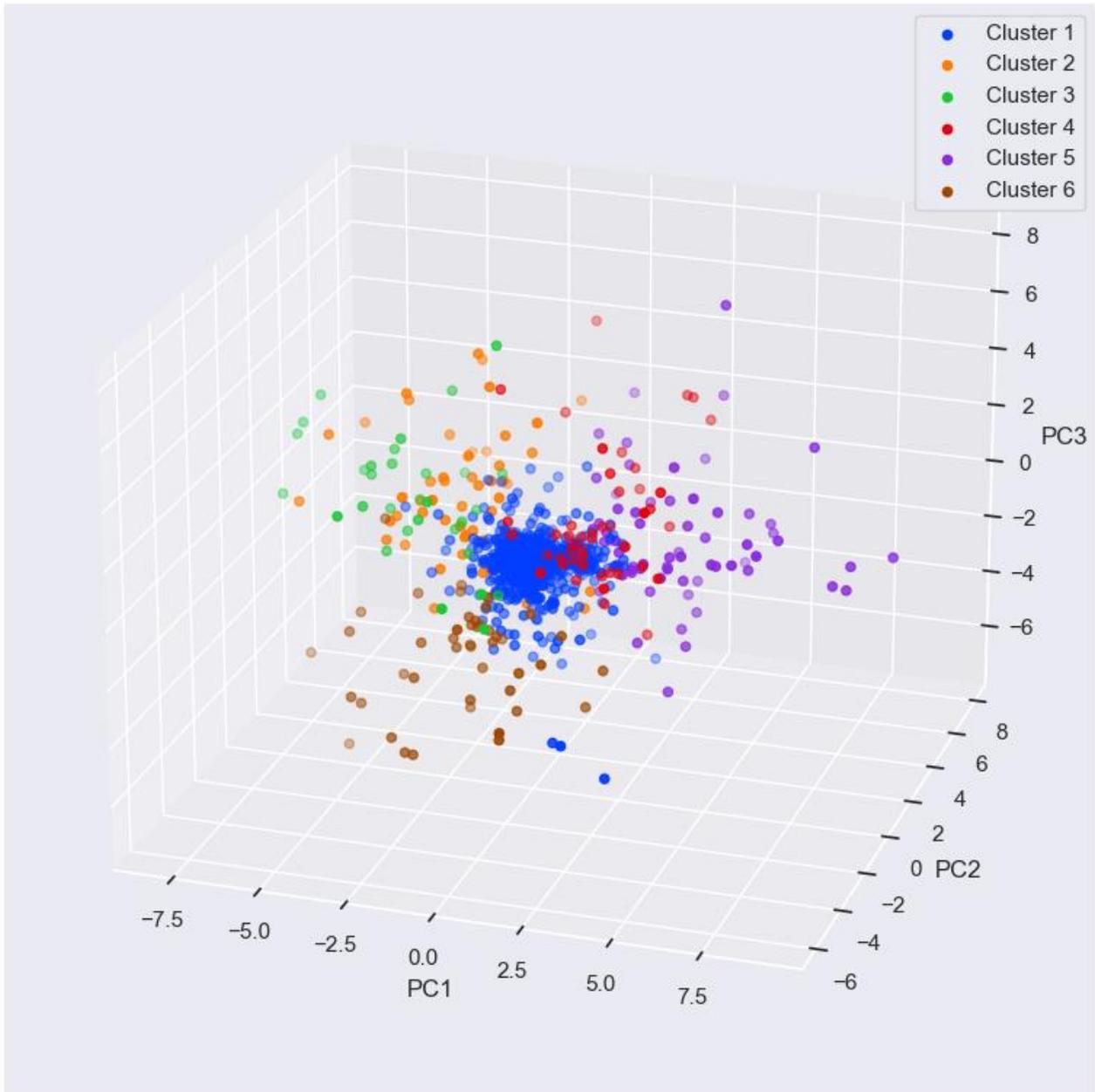

This graph visualizes the classification of keywords from K-Means clustering in a three-dimensional space. Each dot represents a keyword, with its color indicating the respective cluster of the keyword. The three axes capture the first three components from PCA analysis.

**FIGURE 3 Silhouette Analysis for Optimal Number of Clusters**

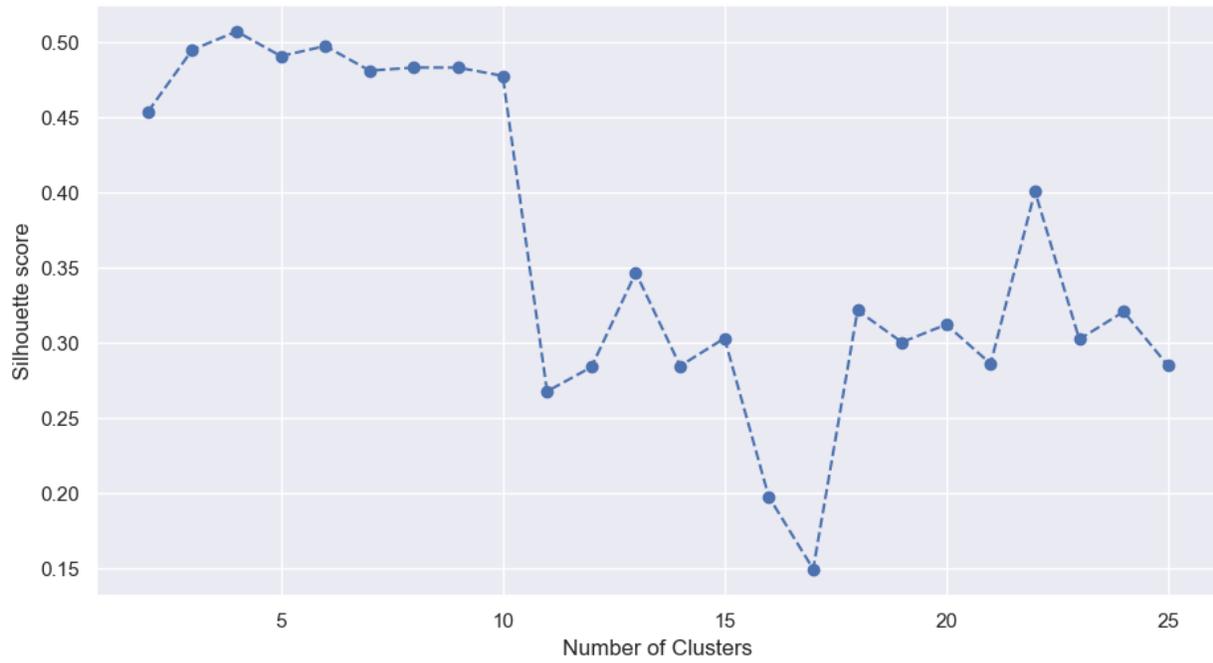

This graph plots the number of clusters and their corresponding Silhouette scores from K-Means clustering.

**FIGURE 4 Trends of HC Disclosures in 10-Ks During the First Year of the SEC Regulation**

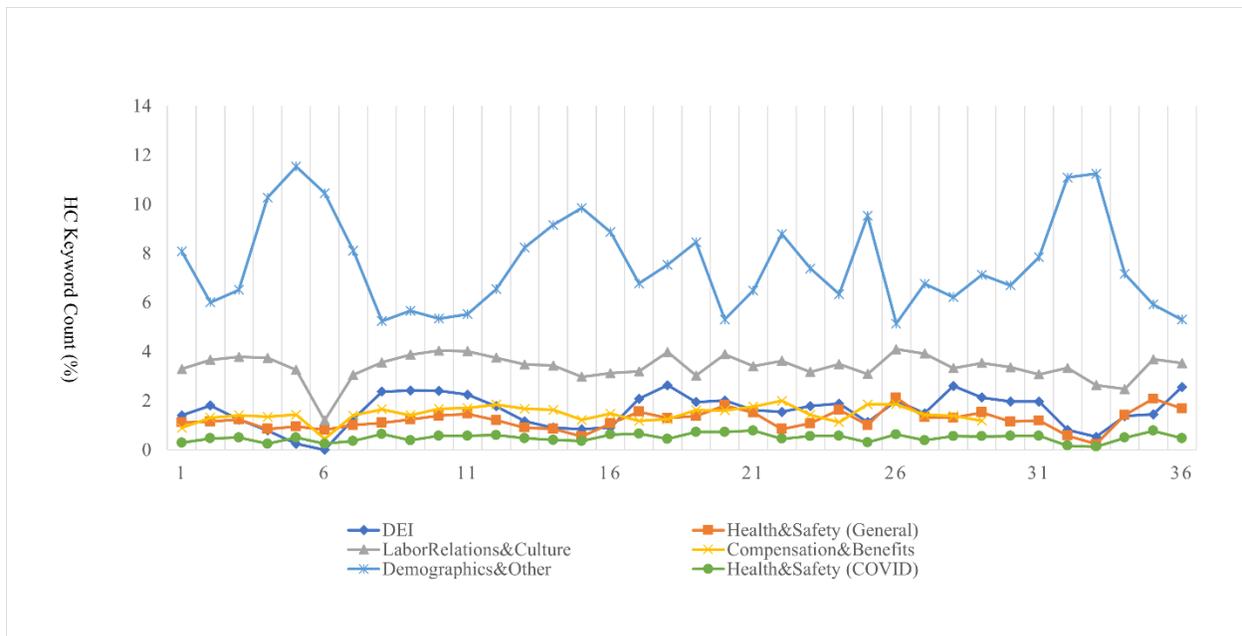

This figure presents the relative HC word count for each category as a percentage of the total word count of HC disclosures in an average 10-K filed during successive 10-day periods from November 2020 to November 2021. On the horizontal axis, "1" represents the first 10 days of this one-year period.

**FIGURE 5 Trend of HC Disclosures in Proxy Statements over 1994-2022**

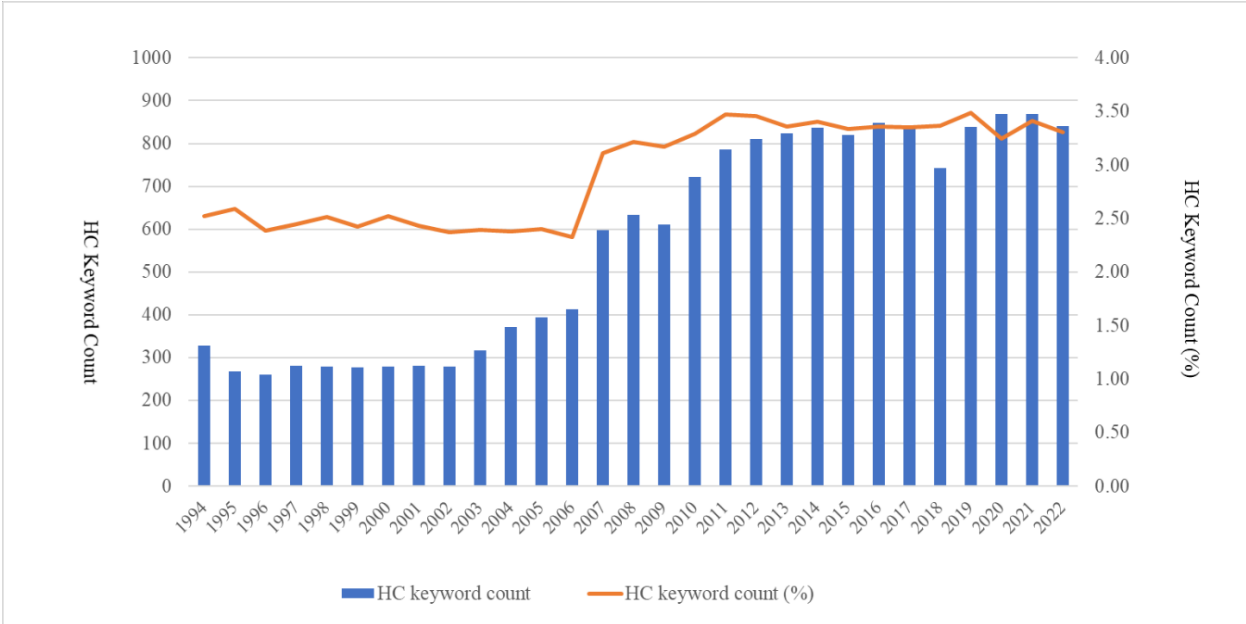

The left vertical axis shows the absolute HC keyword count. The right vertical axis tracks the relative HC keyword count as a percentage of the total word count of an average proxy statement in a given year.

**FIGURE 6 Trend of HC Disclosures in Proxy Statements by Category over 1994-2022**

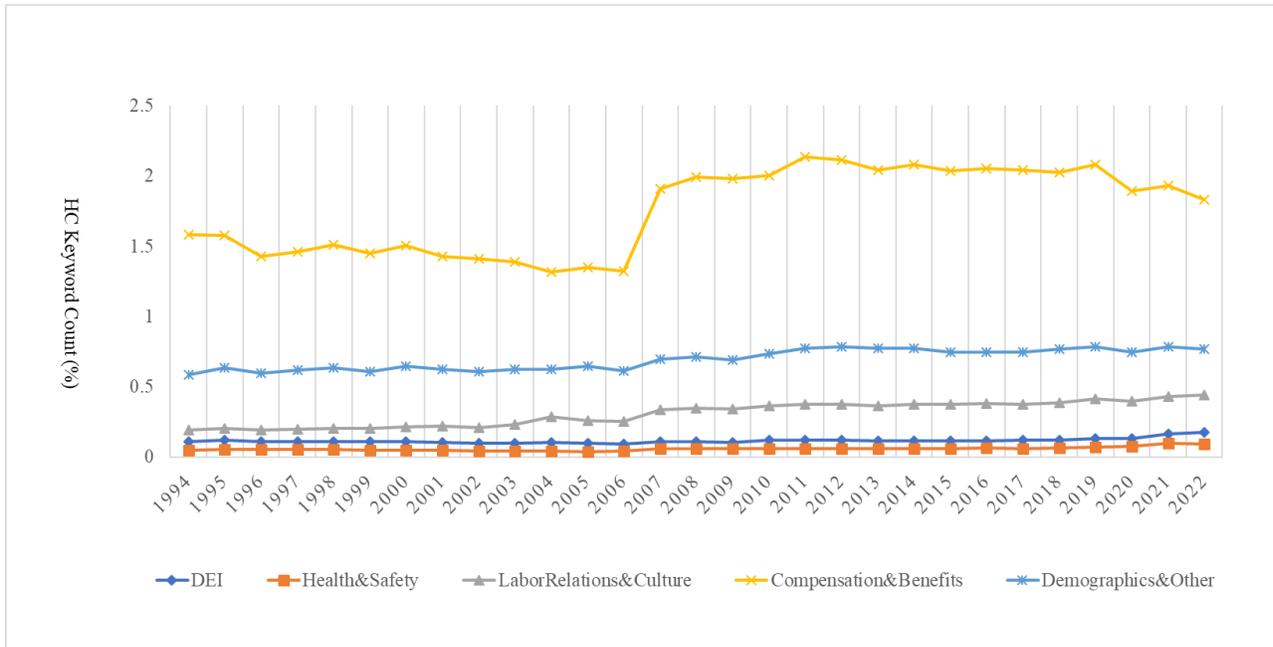

This figure shows the relative HC word count of each category as a percentage of the total word count of an average proxy statement in a given year.

**TABLE 1 Sample Selection**

| | |
|---|---:|
| All 10-K Forms filed during the first year from November 9, 2020 to November 08, 2021 | 7,185 |
| Less: | |
|     Filings from filers not covered by Compustat and CRSP | (3219) |
| Number of 10-Ks in the intersection of EDGAR, Compustat, and CRSP | 3,966 |
| Less: | |
|     Filings for fiscal year 2019 | (5) |
|     Filings from firms having no employees | (3) |
|     Filings that do not contain HC disclosures | (2) |
|     Duplicate filings from the same filer (with the first one kept) | (3) |
| Number of HC disclosures from the same number of unique firms | 3,953 |

This table summarizes the sample selection process for the HC disclosures used to train the *word2vec* model.

**TABLE 2 Seed Word List**

| Category | Seed Words |
|---|---|
| Diversity, Equity, and Inclusion (DEI) | affirmative action, age, allyship, background, bias, black, color, DEI, disability, discriminate, discrimination, discriminatory, diverse, diversity, EEO, equal, equality, equitable, equity, ethnic, ethnicity, fair, female, gender, harassment, hispanic, human right, inclusion, inclusive, indigenous, injustice, justice, LGBT, LGBTQ, minority, nationality, opportunity, origin, pay gap, people of color, race, racial, religion, religious, representation, respect, respectful, sexual orientation, underrepresent, underrepresented, value, woman, women |
| Health and Safety | accident, COVID-19, death, fatality, hazard, health, healthy, illness, incident, injury, mental health, OSHA, pandemic, physical, protect, protection, respiratory, safe, safety, TRIR, vulnerable, well-being, wellness |
| Labor Relations and Culture | absenteeism, appraisal, attract, attraction, attrition, career, career development, child labor, childcare, coaching, code of conduct, collaborative, colleague, collective bargaining, commitment, core value, corrective, culture, develop, development, education, engagement, ethics, evaluate, feedback, forced labor, hire, hiring, labor disruption, mentor, mentoring, mentorship, motivate, organized labor, professional development, promotion, recruit, recruitment, relation, reskill, retain, retention, review, slavery, succeed, success, support, survey, talent, train, training, turnover, union, upskill, whistleblower, work environment, work stoppage |
| Compensation and Benefits | award, benefit, benefits, bereavement, bonus, compensation, contribution, equity award, health care, healthcare, hourly rate, incentive, insurance, maternity, medical, paid leave, parental leave, pay, pension, prescription drug, remuneration, restricted stock, retirement, salaries, salary, sick leave, stock option, vacation, wage |
| Demographics and Others | associate, contract employee, demographic, demographics, employ, employee, full time, full-time, geographical region, geography, headcount, hourly, human capital, human resource, independent contractor, job function, management, part time, part-time, payroll, percent, personnel, region, salaried, sector, segment, skilled, staff, staffing, white-collar, worker, workforce |

# TABLE 3 HC Keyword List

**Panel A:** Diversity, Equity, and Inclusion (DEI) (253 terms)

| Category | Keywords |
|---|---|
| Diversity, Equity, and Inclusion (DEI) | affinity_group, affirmative_action, affirmative_action_plan, african-american, african_american, age, alaska_native, alaskan_native, allyship, american_indian, ancestry, anti-discrimination, anti-harassment, anti-racism, anti-retaliation, asian-american, asian_american, background, belonging, bias, bias-free, bipoc, bisexual, black, brgs, bully, bullying, caucasian, cdo, corporate_equality_index, courtesy, creed, crgs, cross-cultural, cultural_background, dei, dei_council, deib, descent, dib, different_background, different_perspective, disability, disability_equality_index, disabled, discriminate, discrimination, discriminatory, discriminatory_harassment, disparity, diverse, diverse_background, diverse_candidate, diverse_candidate_slate, diverse_perspective, diverse_pool, diverse_slate, diversity, diversity-focused, diversity-related, domestic_partner, edi, eeo, eeoc, eigs, empathy, employment_opportunity, equal-opportunity, equal_employment, equal_employment_opportunity, equal_opportunity, equal_opportunity_employer, equal_pay, equal_treatment, equality, equality_index, equally, equitable, equitable_treatment, equitably, ergs, ethnic, ethnic_background, ethnic_diversity, ethnic_group, ethnic_minority, ethnic_origin, ethnically, ethnically_diverse, ethnicity, fairness, female, gay, gei, gender, gender-equality, gender-neutral, gender_expression, gender_identification, gender_identity, gender_parity, generational, harassment, harassment_prevention, hawaiian, heritage, hispanic, hispanic-serving, historically_black_college, historically_underrepresented, human_right, human_right_campaign, ide, identity, implicit_bias, inclusion, inclusion-focused, inclusion-related, inclusive, inclusive_workplace, inclusively, inclusiveness, inclusivity, indigenous, individuality, inequality, inequity, injustice, intimidation, islander, justice, latino, latinx, lesbian, lgbt+, lgbtq, lgbtq+, lgbtq_equality, lgbtqia, listening_session, living_wage, male, marginalize, marital, marital_status, men, military_status, military_veteran, minority, minority_group, multi-cultural, multicultural, multigenerational, multiracial, naacp, national_origin, nationality, native_american, native_hawaiian, neurodiversity, non-caucasian, non-discrimination, non-discriminatory, non-minorities, non-minority, non-white, nondiscrimination, nondiscriminatory, pacific_islander, pay_equity, pay_gap, pay_parity, people_of_color, poc, political_affiliation, pregnancy, prejudice, protect_veteran, protect_veteran_status, protected_characteristic, protected_class, race, racial, racial_equality, racial_equity, racial_injustice, racial_justice, racial_minority, racially, racially_diverse, racism, reasonable_accommodation, religion, religious, religious_belief, representation, resource_group, respectful_conduct, respectful_workplace, respectfully, retaliation, retribution, self-disclosed, self-expression, self-identification, self-identified, self-identifies, self-identify, self-identifying, sex, sexual, sexual_harassment, sexual_orientation, social_injustice, social_justice, socio-economic, socio-economic_status, socioeconomic, status_protect, stereotype, systemic_racism, traditionally_underrepresented, transgender, treat_fairly, unbiased, unconscious_bias, unconscious_bias_training, under-represented, under-represented_group, under-represented_minority, underprivileged, underrepresentation, underrepresented, underrepresented_community, underrepresented_ethnic, underrepresented_group, underrepresented_minority, underrepresented_population, underserved_community, unique_background, unique_perspective, unlawful_discrimination, urm, urms, values-based, varied_perspective, veteran, veteran-owned, veteran_status, veterans, without_bias, woman, women |

# TABLE 3 HC Keyword List

**Panel B:** Health and Safety (157 general terms and 70 COVID-related terms)

| Category | Keywords |
|---|---|
| Health and Safety (General) | acc, accident, accidental, accidental_death, aes, behavioral_health, biometric_screening, bls, case_rate, clean_supply, cleaning_procedure, cleaning_protocol, cleaning_supply, dart, death, depression, disinfect, disinfectant, disinfecting, disinfection, emergency_response, emotional, emotional_well-being, employee-hours, enhance_cleaning, enhanced_cleaning, ergonomic, ergonomics, fatality, first_aid, first_responder, fitness, fitness_center, fitness_class, flu_shot, flu_vaccination, frequent_cleaning, hand_sanitize_station, hand_sanitizer, handwashing, harm, hazard, hazardous, hazardous_material, health, health-related, health_authority, health_screening, healthful, healthy, healthy_lifestyle, healthy_living, healthy_workplace, hse, hygiene, hygiene_practice, illness, incidence, incidence_rate, incident, incident-free, incident_per, incident_rate, infectious_disease, injury, injury-free, injury_per, injury_rate, labor_statistic, life-saving, lifestyle, lose_time, lose_time_incident_rate, lose_workday, lost-time, lost-time_incident, lost_time, ltcr, lti, ltir, medical_condition, mental, mental_disability, mental_health, mental_well-being, mentally, msha, near-miss, near-misses, occupational, occupational_health, occupational_illness, occupational_injury, occupational_safety, ohi, ohsas, osha, osha_recordable, osha_total_recordable, oshas, personal_protective_equipment, physically, physically_safe, ppe, precautionary_measure, prevent_injury, preventative_measure, protection_equipment, protective_equipment, protective_gear, provide_personal_protective, public_health, public_health_authority, qhse, recordable_incident, recordable_incident_rate, recordable_injury, recordable_injury_rate, rir, safe, safe_working, safely, safer, safety, safety-first, safety-focused, safety-related, safety_protocol, sanitation, sanitation_procedure, sanitization, sanitize, sanitizer, sanitizers, sanitizing, sanitizing_procedure, serious_injury, sif, tir, total_recordable_incident_rate, trir, unsafe, unsafe_condition, ventilation, vpp, welfare, well-being, wellness, work-related_fatality, work-related_injury, working_condition, workplace_injury, workspace, zero-incident, zero_accident, zero_incident, zero_injury |
| Health and Safety (COVID) | cdc, cdc_guideline, coronavirus, coronavirus_disease, covid, covid-19, covid-related, daily_temperature_check, essential_worker, face-masks, face_cover, face_covering, face_mask, facemasks, international_non-essential_travel, mask, mask-wearing, mask_requirement, mask_wearing, masking, non-essential_travel, novel_coronavirus, pandemic, pandemic-related, physical_distance, physical_distancing, proper_social_distancing, quarantine, quarantined, quarantining, remote-work, remote-working, remote_work, remote_work_arrangement, remote_work_environment, remote_working, reopening, respirator, self-health, self-quarantine, self-screening, shelter-in-place, social-distancing, social_distance, social_distancing, social_distancing_guideline, stay-at-home, stay-at-home_order, telecommute, telecommuting, teleworking, temperature-taking, temperature_check, temperature_screen, temperature_screening, test_positive, travel_restriction, vaccinate, vaccination, vaccine, virus, virus_exposure, wear_face_covering, wear_mask, work-at-home, work-from-home, work-from-home_arrangement, work-from-home_policy, work_remotely, working-from-home |

# TABLE 3 HC Keyword List

**Panel C:** Labor Relations and Culture (362 terms)

| Category | Keywords |
|---|---|
| Labor Relations and Culture | absenteeism, abuse, abusive, advancement, afl-cio, alpa, anonymous, anonymous_feedback, anonymous_hotline, anonymous_survey, anonymously, anti-bribery, anti-corruption, anticorruption, applicant, at-will, attendance_policy, attract, attracting, attraction, attracts, attrition, attrition_rate, bargain_unit, bargaining_group, bargaining_unit, best_employer, bribery, candid_feedback, candidate_pool, candidate_slate, candor, career, career-enhancing, career-oriented, career_advancement, career_development, career_growth, career_mobility, career_path, career_progression, charitable_cause, charitable_contribution, charitable_donation, charitable_foundation, charitable_gift, charitable_give, charitable_matching, charitable_organization, charity, child_labor, chro, co-worker, coaching, code_of_conduct, cohort, collaboration, collaborative, colleague, collective-bargaining_agreement, collective_bargaining, collective_bargaining_agreement, collective_bargaining_arrangement, collective_bargaining_unit, collegial, collegiality, commitment, community_involvement, community_outreach, commute, compassion, competency-based, compliance_hotline, confidential_hotline, confidential_reporting, confidentially, consultation, continual_learning, continuous_learn, continuous_learning, core_principle, core_value, corporate_philanthropy, corrupt, corruption, counseling, counseling_session, counselling, courage, courageous, coworkers, credential, cross-training, cultivate, cultural, cultural_awareness, cultural_fit, culture, customized_corporate_training, cwa, decency, dependent_care, direct_report, disaster_relief, eap, early-career, early_career, education, elearning, emerge_leader, emotional_intelligence, empathetic, employee-centric, employee-driven, employee-focused, employee-friendly, employee-led, employee-led_group, employee_value_proposition, employer_brand, employment-related, employment-related_work_stoppage, employment_decision, empowered, empowerment, engagement, engagement_score, engagement_survey, engaging, enjoyable_workplace, ethic, ethic_hotline, ethical, ethical_conduct, ethically, ethics, ethicsline, evaluation_process, executive-sponsored, exit_interview, external_hire, fair_treatment, fairly_compensate, family, fast-paced, flexible_schedule, forced_labor, fraud, fundraise, fundraising, furlough, furloughed, grievance, high-performance_culture, high_ethical_standard, highly_collaborative, hire, hiring, hiring_practice, honest, honestly, human_trafficking, humble, humility, iam, ibew, ibt, ilo, impasse, inappropriate_behavior, inspirational, inspiring, integrity, internal_candidate, internal_mobility, internal_promotion, international_brotherhood, international_union, internship, interview, interview_panel, interviewer, interviewing, involuntary_turnover, iuoe, job-related, job-specific, job_assignment, job_description, job_duty, job_fair, job_opening, job_satisfaction, labor-related, labor-related_work_stoppage, labor_disruption, labor_law, labor_relation, labour, laundering, lay-offs, layoff, leadership, learning, learning_management_system, like-minded, material_work_stoppage, mentor, mentoring, mentorship, mentorships, meritocracy, mid-career, moral, motivate, motto, new-hire, new_hire, newly_hire, non-profit_organization, non-unionized, nonprofit_organization, on-board, on-boarding, on-demand_learning, on-the-job, on-the-job_learning, on-the-job_training, onboarding, onboarding_process, one-on-one, online_learning, opeiu, open-door_policy, open_communication, open_dialogue, open_position, organized_labor, orientation, pafca, people-centric, people-first, performance_appraisal, performance_evaluation, personality, philosophy, post-employment, potential_successor, professional_development, professionalism, professionally, promote-from-within, promotion, pulse_survey, questionnaire, recruit, recruiter, recruiting, recruitment, reduction-in-force, reductions-in-force, renegotiation, representative_body, reprisal, resignation, reskill, reskilling, results-driven, retain, retain_qualified, retain_talented, retain_top, retaining, retains, retention, retention_rate, retentive, review_process, reward, rotational_assignment, self-assessment, self-development, self-directed_learning, self-paced_learning, significant_work_stoppage, skill-based, skill-building, skills-based, skillset, |

| Category | Keywords |
|---|---|
| | skillsets, slavery, soft_skill, solicit_feedback, succession, succession_plan, succession_planning, succession_planning_process, supervisor, survey, talent_acquisition, talent_attraction, talent_mobility, talent_pipeline, talent_pool, team-based, team-building, team-oriented, tenet, tenure, tenured, termination, top-performing, top_talent, town_hall_meeting, townhall, townhalls, trade_union, train, trained, trainee, trainer, training, turnover, turnover_rate, tutor, tutoring, uaw, umwa, unethical, unethical_behavior, union, union_contract, unionization, unionize, unionized, united_way, upskill, upskilling, uwua, values-based_culture, virtual_classroom, virtual_town_hall, voluntary_attrition, voluntary_attrition_rate, voluntary_separation, voluntary_turnover, voluntary_turnover_rate, volunteer, volunteer_activity, volunteer_hour, volunteering, vto, whistleblower, whistleblower_hotline, whistleblower_policy, work-life, work-life_balance, work_environment, work_stoppage, worker_council, working_environment, workplace |

# TABLE 3 HC Keyword List

**Panel D:** Compensation and Benefits (283 terms)

| Category | Keywords |
|---|---|
| Compensation and Benefits | 401k, absence, accident_insurance, adoption_assistance, annual_bonus, annual_cash, annual_incentive, assistance_fund, assistance_program, autism, base_salary, base_wage, benefit, benefit_package, benefits, bereavement, bereavement_leave, bonus, bonus_award, bonus_payment, broad-based_equity, broad-based_equity_award, broad-based_stock_grant, broad-based_stock_incentive, care_provider, caregiver, caregiver_leave, caregiving, cash-based, cash-based_compensation_award, cash-based_performance_bonus, cash_bonus, cash_incentive, cash_incentive_plan, child-care, child_care, childcare, co-pays, commission, commission-based, commuter_benefit, company-matched, company-paid, company-paid_benefit, company-provided, company-subsidized, compensate_fairly, compensation, compensation_package, compensation_philosophy, competitive_base_salary, competitive_benefit_package, competitive_compensation, competitive_compensation_package, competitive_pay, competitive_salary, competitive_wage, complimentary, comprehensive_benefit_package, confidential_counseling, cost-sharing, critical_illness, daycare, deferred_compensation, define_contribution, defined_benefit, defined_contribution, dental, dental_insurance, dependent, dependent_care_flexible_spending, disability_coverage, disability_insurance, discretionary_bonus, dismemberment, education_assistance, education_reimbursement, educational_assistance, elder_care, eldercare, employee_stock_purchase, employer-paid, employer-sponsored, employer_contribution, employer_match, employment_package, equity-based, equity-based_award, equity-based_compensation, equity-based_compensation_award, equity-based_grant, equity-based_incentive, equity_award, equity_grant, equity_ownership, esop, espp, eto, fair_wage, family-friendly, family_care, family_leave, family_member, financial_assistance, financial_hardship, financial_well-being, fitness_reimbursement, flexible_spending, flexible_spending_account, flexible_time-off, flexible_work_schedule, flextime, generous_pay_time, health_care, health_saving, healthcare, high-performing, high_performer, holiday, hospital_indemnity, hourly_basis, hourly_rate, hourly_wage, hsa, identity_theft, identity_theft_insurance, immediate_family, incentive, incentive-based, incentive_award, incentive_bonus, incentive_compensation, incentive_plan, incentivize, incentivized, incentivizes, incentivizing, income_protection, industry_peer, insurance, insurance_coverage, juneteenth, jury, ksop, life_insurance, long-term_disability, long-term_disability_insurance, long-term_incentive, long_term_disability, loved_one, ltip, market-competitive_compensation, market-competitive_pay, market-competitive_salary, market_median, match_contribution, matching_contribution, maternal, maternity, maternity_leave, medical_coverage, medical_insurance, merit-based, military_leave, minimum_wage, new_parent, overtime, paid-time, paid-time-off, paid_leave, paid_time, parent, parental, parental_leave, parental_leave_policy, parenting, paternal, paternity, pay, pay-for-performance, pay-for-performance_philosophy, pay_holiday, pay_parental, pay_parental_leave, pay_time, pay_time-off, pay_vacation, paycheck, payroll, payroll_deduction, pension, pension_plan, performance-based, performance-based_bonus, performance-based_cash, performance-driven, perk, perquisite, pet_insurance, physical_therapy, post-retirement, prescription_drug, prescription_drug_benefit, profit-sharing, profit_sharing, pto, reimbursement, relief_fund, remuneration, restrict_stock, restrict_stock_unit, restricted_stock, restricted_stock_unit, retiree, retirement, retirement_benefit, retirement_plan, retirement_planning, retirement_saving, retirement_saving_plan, reward_package, rsus, sabbatical, salaries, salary, salary_reduction, sale_commission, save_plan, saving_account, saving_plan, scholarship, severance, share-based, share-based_compensation, short-term_disability, short-term_incentive, sick_day, sick_leave, sick_time, smoke_cessation, spending_account, spouse, stipend, stock-based, stock-based_award, stock-based_compensation, |

| Category | Keywords |
|---|---|
| | stock-based_compensation_award, stock_award, stock_grant, stock_option, stock_option_grant, stock_ownership, stock_ownership_plan, stock_purchase_plan, student_loan, supplemental_life, surrogacy_assistance, tax-advantaged, tax-deferred_basis, time-off, top_performer, total_compensation_package, total_reward, total_reward_package, tuition, tuition_assistance, tuition_reimbursement, tuition_reimbursement_program, unpaid_leaf, unpaid_leave, vacation, vacation_time, valuable_fringe, variable_compensation, variable_pay, vest, vest_condition, vested, vesting, vision_coverage, vision_insurance, wage, welfare_benefit |

# TABLE 3 HC Keyword List

**Panel E:** Demographics and Others (160 terms)

| Category | Keywords |
|---|---|
| Demographics and Others | administrator, advanced_degree, agent, apprentice, assistant_manager, associate, average_tenure, bachelor, back-office, clerk, co-op, college_graduate, college_student, consultant, contingent_worker, contract_employee, contract_worker, contractor, crew, crew_member, crewmember, crewmembers, demographic, demographics, district_manager, doctorate, electrical_worker, employ, employ_approximately, employed, employee, employee-related, engineer, entire_workforce, entry-level, entry_level, executive-level, executive_leadership, executive_officer, fixed-term, front-line, front-office, frontline, fte, full-time, full-time_basis, full-time_equivalent, full_time, full_time_equivalent, fulltime, general_manager, graduate_degree, headcount, high-caliber, high_caliber, high_performing, high_potential, highly-qualified, highly-skilled, highly_educate, highly_qualified, highly_qualify, highly_skilled, highly_talented, highly_trained, hour_per_week, hourly, hourly_team_member, hris, human-capital, human_capital, human_capital_management, human_resource, independent_agent, independent_consultant, independent_contractor, job_function, labor_cost, labor_force, labor_market, laborer, long-tenured, management-level, management_trainee, manager, manager-level, master_degree, mid-level_manager, motivated, name_executive_officer, non-employee, non-executive, non-management, non-represented, non-salaried, non-union, outside_consultant, part-time, part-time_basis, part_time, participation_rate, people, per_hour, per_week, personally, personnel, pre-employment, professional, qualified, qualified_applicant, qualified_individual, qualified_personnel, recent_graduate, salaried, sale_force, sale_representative, salesperson, senior-level, senior_executive, senior_leader, senior_leadership, senior_leadership_team, senior_management, senior_management_team, senior_manager, seniority, shift_schedule, skill, skilled, staff, staff_member, staffed, staffing, staffing_level, stagger_shift, stem-related, subcontractor, talent, talented_individual, team_member, teammate, teammates, technician, technicians, temporary_worker, unemployment, unskilled, unskilled_labor, vacancy, vocational, well-qualified, well-trained, work-related, work_schedule, workday, worker, workforce, workforce_demographic, workload, workweek |

# TABLE 4 Definitions of Acronyms in Keyword List

| Acronym | Full Spelling |
| --- | --- |
| ACC | American Chemistry Council |
| AES | Asbestos Environment and Safety |
| AFL-CIO | American Federation of Labor and Congress of Industrial Organizations |
| ALPA | Air Line Pilots Association |
| BIPOC | Black, Indigenous and People of Color |
| BLS | Bureau of Labor Statistics |
| BRGS | Best Employee Resource Groups |
| CDC | Centers for Disease Control and Prevention |
| CDO | Chief Diversity Officer |
| CHRO | Chief Human Resources Officer |
| CRGS | Colleague Resource Groups |
| CWA | Communications Workers of America |
| DART | Days Away Restricted or Transferred |
| DEI | Diversity, Equity and Inclusion |
| DEIB | Diversity, Equity, Inclusion, and Belonging |
| DIB | Diversity, Inclusion, and Belonging |
| EAP | Employee Assistance Program |
| EDI | Equality, Diversity and Inclusion |
| EEO | Equal Employment Opportunity |
| EEOC | Equal Employment Opportunity Commission |
| EIGS | Employee Inclusion Groups |
| ERGS | Employee Resource Groups |
| ESOP | Employee Stock Ownership Plan |
| ESPP | Employee Stock Purchase Plan |
| ETO | Earned Time Off |
| FTE | Full-time Equivalency |
| GEI | Gender-Equality Index |
| HRIS | Human Resources Information System |
| HSA | Health Savings Accounts |
| HSE | Health, Safety, and Environment |
| IAM | International Association of Machinists and Aerospace Workers |
| IBEW | International Brotherhood of Electrical Workers |
| IBT | International Brotherhood of Teamsters |
| IDE | Inclusion, Diversity & Equity |
| ILO | International Labour Organization |
| IUOE | International Union of Operating Engineers |
| KSOP | A retirement plan that combines 401(K) and ESOP |

| Acronym | Full Spelling |
|---|---|
| LTCR | Lost Time Case Rate |
| LTI | Lost Time Injury |
| LTIP | Long Term Incentive Plan |
| LTIR | Lost Time Incident Rate |
| MSHA | Mine Safety and Health Administration |
| NAACP | National Association for the Advancement of Colored People |
| OHI | Other Health Impairment |
| OHSAS | Occupational Health and Safety Assessment Series |
| OPEIU | Office and Professional Employees International Union |
| OSHA | Occupational Safety and Health Administration |
| PAFCA | Professional Airline Flight Control Association |
| POC | People of Color |
| PPE | Personal Protective Equipment |
| PTO | Paid Time Off |
| QHSE | Quality, Health, Safety & Environment |
| RIR | Reportable Incidence Rate |
| RSUS | Restricted Stock Units |
| SIF | Serious Injuries and Fatalities |
| TIR | Total Incident Rate |
| TRIR | Total Recordable Incident Rate |
| UAW | United Auto Workers |
| UMWA | United Mine Workers of America |
| URM | Underrepresented Minority |
| URMS | Underrepresented Minorities |
| UWUA | Utility Workers Union of America |
| VPP | Voluntary Protection Programs |
| VTO | Volunteer Time Off |

**TABLE 5 HC Disclosure Topics by Fama-French 12-Industry Classification**

| Fama-French 12 Industries | DEI | Health and Safety (General) | Health and Safety (COVID) | Labor Relations and Culture | Compensations and Benefits | Demographics and Others |
|---|---|---|---|---|---|---|
| Telephone and Television Transmission | 17.49% | 7.16% | 3.81% | 24.85% | 11.54% | 35.16% |
| Finance | 15.54% | 7.70% | 4.26% | 25.17% | 13.65% | 33.69% |
| Healthcare, Medical Equipment, and Drugs | 12.43% | 8.21% | 4.12% | 24.73% | 14.28% | 36.23% |
| Business Equipment -- Computers, Software, and Electronic Equipment | 17.16% | 8.29% | 4.06% | 26.32% | 10.99% | 33.18% |
| Wholesale, Retail, and Some Services (Laundries, Repair Shops) | 14.30% | 10.01% | 4.60% | 24.50% | 12.27% | 34.33% |
| Other -- Mines, Constr, BldMt, Trans, Hotels, Bus Serv, Entertainment | 13.77% | 11.31% | 4.01% | 26.67% | 9.53% | 34.72% |
| Consumer NonDurables -- Food, Tobacco, Textiles, Apparel, Leather, Toys | 16.76% | 11.81% | 4.29% | 24.77% | 10.27% | 32.10% |
| Consumer Durables -- Cars, TV's, Furniture, Household Appliances | 14.59% | 12.07% | 4.31% | 25.96% | 9.32% | 33.75% |
| Chemicals and Allied Products | 13.64% | 14.20% | 3.70% | 25.23% | 8.81% | 34.43% |
| Manufacturing -- Machinery, Trucks, Planes, Off Furn, Paper, Com Printing | 13.61% | 14.71% | 4.55% | 25.24% | 8.90% | 33.00% |
| Utilities | 15.32% | 14.89% | 3.80% | 25.19% | 9.67% | 31.13% |
| Oil, Gas, and Coal Extraction and Products | 12.81% | 15.39% | 3.29% | 25.47% | 10.43% | 32.60% |

**APPENDIX   Data and Code**

[ISYS-2023-023_HC_Disclosures_Data_and_Code.zip](ISYS-2023-023_HC_Disclosures_Data_and_Code.zip)